\documentclass{article}

\usepackage[preprint,nonatbib]{neurips_2021}

\usepackage[utf8]{inputenc} 
\usepackage[T1]{fontenc}    
\usepackage{hyperref}       
\usepackage{url}            
\usepackage{booktabs}       
\usepackage{amsfonts}       
\usepackage{nicefrac}       
\usepackage{microtype}      
\usepackage{amsmath}
\usepackage{floatrow}

\DeclareMathOperator*{\argmin}{arg\,min}

\newcommand{\afterfigure}{\vspace{-1em}}

\usepackage{xcolor}
\usepackage{graphicx}
\newcommand{\bb}{{\bf b}}
\newcommand{\bba}{{\bf A}}
\newcommand{\bbi}{{\bf I}}
\newcommand{\bbm}{{\bf M}}
\newcommand{\bbt}{{\bf T}}
\newcommand{\bbw}{{\bf W}}
\newcommand{\Loss}{{\mathcal{L}}}

\newif\ifdraft
\draftfalse

\ifdraft
    \newcommand{\kac}[1]{{\color{magenta}\textbf{Kfir:} #1}}
    \newcommand{\jhc}[1]{{\color{red}\textbf{Junfeng:} #1}}
    \newcommand{\mrc}[1]{{\color{orange}\textbf{Miki:} #1}}
    \newcommand{\ygc}[1]{{\color{green}\textbf{Yossi:} #1}}
    \newcommand{\imc}[1]{{\color{cyan}\textbf{Inbar:} #1}}
    \newcommand{\dejc}[1]{{\color{brown}\textbf{David:} #1}}
    \newcommand{\ypc}[1]{{\color{blue}\textbf{Yael:} #1}}
    \newcommand{\kjk}[1]{{\color{purple}\textbf{Kai:} #1}}

\else
    \newcommand{\kac}[1]{}
    \newcommand{\jhc}[1]{}
    \newcommand{\mrc}[1]{}
    \newcommand{\ygc}[1]{}
    \newcommand{\imc}[1]{}
    \newcommand{\dejc}[1]{}
    \newcommand{\ypc}[1]{}
    \newcommand{\kjk}[1]{}

\fi

\title{Deep Saliency Prior for Reducing Visual Distraction}

\author{%

Kfir Aberman\thanks{Denotes equal contribution} \;\;\;\;\;
Junfeng He$^*$\;\;\;\;\;
  Yossi Gandelsman \\
  \textbf{Inbar Mosseri \;\;\;\;\;
  David E. Jacobs \;\;\;\;\;
    Kai Kohlhoff} \\

  \textbf{Yael Pritch \; \; \;\;\;
  Michael Rubinstein} \\

  Google Research 
}

\begin{document}

\maketitle

\begin{abstract}
Using only a model that was trained to predict where people look at images, and no additional training data, we can produce a range of powerful editing effects for reducing distraction in images. Given an image and a mask specifying the region to edit, we backpropagate through a state-of-the-art saliency model to parameterize a differentiable editing operator, such that the saliency within the masked region is reduced. We demonstrate several operators, including: a recoloring operator, which learns to apply a color transform that camouflages and blends distractors into their surroundings; a warping operator, which warps less salient image regions to cover distractors, gradually collapsing objects into themselves and effectively removing them (an effect akin to inpainting); a GAN operator, which uses a semantic prior to fully replace image regions with plausible, less salient alternatives. The resulting effects are consistent with cognitive research on the human visual system (e.g., since color mismatch is salient, the recoloring operator learns to harmonize objects' colors with their surrounding to reduce their saliency), and, importantly, are all achieved solely through the guidance of the pretrained saliency model, with no additional supervision. We present results on a variety of natural images and conduct a perceptual study to evaluate and validate the changes in viewers' eye-gaze between the original images and our edited results.

\end{abstract}

\section{Introduction}

Studying and modeling human attention -- how and where people look at images -- has been widely researched and explored. In the deep learning era, saliency models trained on eye-gaze data are now able to predict human visual attention to high accuracy. However, while the research community has so far focused on developing models for \emph{predicting} where people look, almost no attention has been given to utilizing the knowledge embedded in such recent, deep saliency models to actually \emph{drive and direct} editing of images and videos, so as to tweak the attention drawn to different regions in them. A few recent attempts~\cite{gatys2017guiding, mejjati2020look} have focused on subtle effects designed to make minimal modifications to the image, and are therefore limited in their ability to make meaningful changes to visual attention.

In this paper, we leverage deep saliency models to drive dramatic, but still realistic, edits, which can significantly change an observer's attention to different regions in an image. Such capability can have important applications, for example in photography, where pictures we take often contain objects that distract from the main subject(s) we want to portray, or in video conferencing, where clutter in the background of a room or an office may distract from the main speaker participating in the call.

We ask: using a differentiable saliency model as a guide, what types of editing effects can be achieved?  How would those effects affect viewers' attention in practice when looking at the images?  Our focus in this paper is on \emph{decreasing attention} for the purpose of reducing visual distraction, but we also demonstrate some results for \emph{increasing} attention drawn to image regions in Section~\ref{sec:results} (Fig.~\ref{fig:gan}).

\begin{figure}
\centering
\includegraphics[width=\columnwidth]{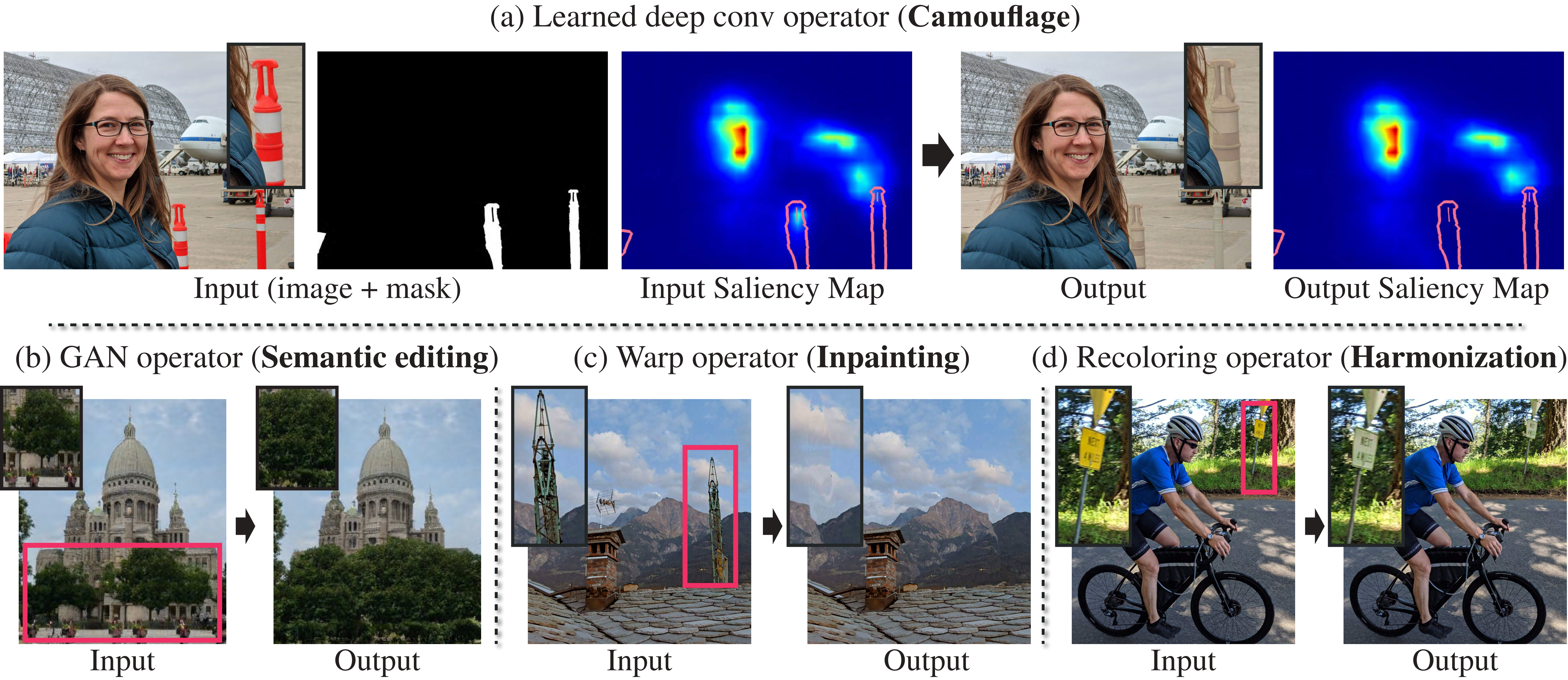}
\caption{Given an input image and a mask of the region(s) to edit (top row, left), our method back-propagates through a visual saliency prediction model to solve for an image such that the saliency level in the region of interest is modified (top row, right). We explore a set of differentiable operators, the parameters of which are all guided by the saliency model, resulting in a variety of effects such as (a) camouflaging (b) semantic editing (c) inpainting, and (d) color harmonization. \afterfigure}

\label{fig:teaser}
\end{figure}

To this end, we develop an optimization framework for guiding visual attention in images using a differentiable, predictive saliency model. Our method employs a state-of-the-art deep saliency model~\cite{jia2020eml}, pre-trained on large-scale saliency data~\cite{jiang2015salicon}. Given an input image and a distractor mask, we backpropagate through the saliency model -- effectively using it as a \emph{prior} -- to parameterize an editing operator, such that the saliency within the masked region is reduced (Fig.~\ref{fig:teaser}). The space of appropriate operators in such a framework is, however, not unbounded. The problem lies in the saliency predictor---as with many deep learning models, the parametric space of saliency predictors is sparse and prone to failure if out-of-distribution samples are produced in unconstrained manner (Figure~\ref{fig:adversarial_attack}). Using a careful selection of operators and priors, we show that natural and realistic editing can be achieved via gradient descent on a single objective function.

We experiment with several differentiable operators: two standard image editing operations (whose parameters are learned through the saliency model), namely recolorization and image warping (shift); and two learned operators (we do not define the editing operation explicitly), namely a multi-layer convolution filter, and a generative model (GAN). With those operators, our framework is able to produce a variety of powerful effects, including recoloring, inpainting, detail/tone attenuation, camouflage, object editing or insertion, and facial attribute editing (Figure~\ref{fig:teaser}). Importantly, all these effects are driven solely by the single, pretrained saliency model, without any additional supervision or training. Note that our goal is not to compete with dedicated methods for producing each effect, but rather to demonstrate how multiple such editing operations can be guided by the knowledge embedded within deep saliency models, all within a single framework.

We demonstrate our approach on a variety of natural images, and conduct a perceptual study to validate the changes in real human eye-gaze between the original images and our edited results. Our experiments and user studies show that the produced image edits: a) effectively reduce the visual attention drawn to the specified regions, b) maintain well the overall realism of the images, and c) are significantly more preferred by users over more subtle saliency-driven editing effects that were proposed before.

\section{Related Work}

\subsection{Human visual attention and saliency prediction models} 
\label{sec:attention_saliency}
Existing research on human visual attention has demonstrated that our attention is attracted to visually salient stimuli, i.e., a region sufficiently different from its surroundings, in terms of color, intensity, size, spatial frequency, orientation, shape, etc. \cite{frintrop2010computational,Itti2007scholarpedia,wolfe2004attributes,wolfe2017five}. Moreover, studies were shown that human visual attention is drawn by particular objects like faces, texts \cite{cerf2009faces}, and emotion eliciting stimuli \cite{borji2019saliency,fan2018emotional}, which are important for our survival.

\begin{figure}
\centering
\includegraphics[width=\columnwidth]{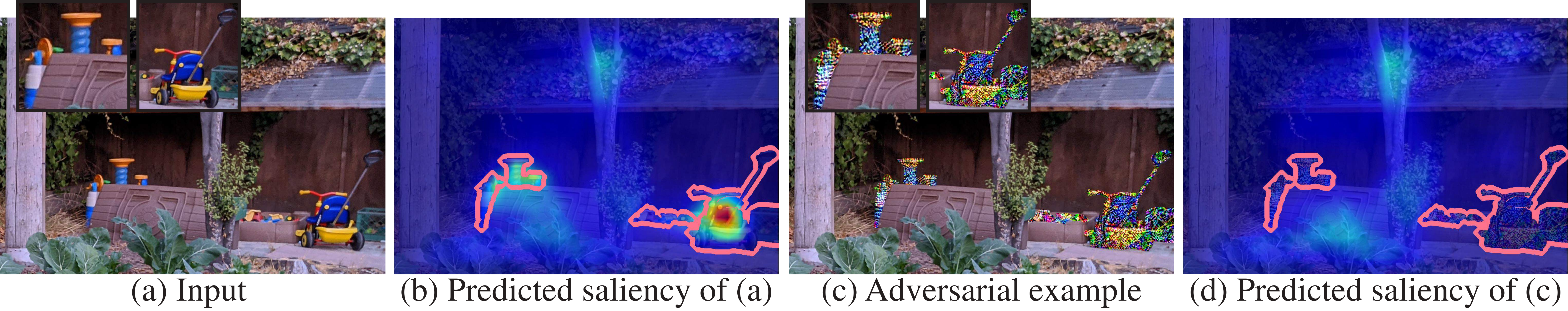}
\caption{
An adversarial example of saliency models. Given an input image (a) with a predicted saliency (b), additive noise is applied to the image and optimized to reduce the saliency of image regions that were previously salient. However, the output (c) still exhibits salient regions which are interpreted as non-salient by the model (d). \afterfigure}
\label{fig:adversarial_attack}
\end{figure}

Saliency prediction models aim at predicting which areas in an image will be salient to human attention and attract eye fixation. Early works in saliency prediction usually define saliency through a set of  hand crafted features such as color difference. contrast, intensities, etc. \cite{itti2001computational,koch1987shifts}. Recent works \cite{huang2015salicon,jia2020eml, kummerer2014deep,kummerer2017understanding} leverage the power of deep neural networks and are often trained/fine-tuned on large scale gaze data set \cite{MITSaliencyBenchmark, jiang2015salicon}.  Deep neural networks based saliency prediction models often perform quite well, with predicted salient regions matching human gaze ground truths \cite{jia2020eml, kummerer2017understanding}.
A more thorough review on saliency prediction models can be found in \cite{borji2019saliency,frintrop2010computational}.

\subsection{Saliency Driven Image Manipulation}
Saliency prediction models have been applied to various applications such as image/video compression \cite{patel2021saliency}, quality assessment \cite{zhang2015application}, visualization \cite{bylinskii2017learning}, and image captioning \cite{cornia2018paying}. Specifically, saliency prediction models are shown to be helpful for image editing tasks \cite{ghosh2019saliency,gu2014automatic,wong2011saliency}, e.g., to enhance contrast \cite{gu2014automatic}, improve aesthetics \cite{wong2011saliency}, and enhance details \cite{ghosh2019saliency}. 

There are some early works on using saliency models to guide human attention~\cite{hagiwara2011saliency, mateescu2014attention, mechrez2019saliency}, however, they either do not use deep saliency models, or only use it as an extra input. Only recently, a few approaches \cite{chen2019guide,gatys2017guiding,mejjati2020look} suggested using deep saliency prediction models in the loss function with back propagation to help retarget visual attention.
Gatys et al.~\cite{gatys2017guiding} use a neural network that receives an image and a target saliency map, and generates an image satisfying that map. 
Chen et al.~\cite{chen2019guide} use a similar architecture with an additional cyclic loss to stabilize the training procedure and reduce artifacts. 
Both of these approaches applies an adversarial and perceptual losses to the output, which strictly restricts its deviation from the original content of the region, resulting in a subtle and narrow effect. 
Recently, Mejjati et al.~\cite{mejjati2020look} proposed a neural network to predict a set of parameters that are applied to the image via a set of pre-defined operators, imitating the subtle changes that professional editors apply to images in order to retarget attention while maintaining fidelity to the original image. 
All of the previous approaches have been able to show only narrow and subtle effects, producing a single result with minimal changes to the visual saliency.  
In contrast, our approach utilizes the modeled perception of the saliency detector to a full extent, proposing a set of effects that are more dramatic and effective in guiding the visual attention. Moreover, unlike previous approaches that require large-scale datasets, our approach is simple and doesn't involve any training data, or training sessions, and is simple to tune. It requires a single saliency model that was pre-trained on high-quality eye-gaze tracking data.

\section{Method}
\label{sec:method}
Given an input image $\bbi$ and a region of interest $\bbm$, our objective is to manipulate the content of $\bbi$ such that the attention drawn to region $\bbm$ is modified while keeping high-fidelity to the original image in other areas. Our approach is to follow the guidance of a saliency prediction model \cite{jia2020eml}\footnote{For all the experiments in this paper we use the saliency prediction model of~\cite{jia2020eml}, with minor modifications that are described in the supplementary material.} that was pretrained to identify attention grabbing regions based on saliency data \cite{jiang2015salicon}. Formally, we seek to find an image $\tilde{\bbi}$ that solves the following, two-term optimization problem:
\begin{equation}
\label{main_naive_obj}
\argmin_{\tilde{\bbi}} \; \Loss_{\text{sal}}\left(\tilde{\bbi}\right) + \beta \Loss_{\text{sim}}\left(\tilde{\bbi}\right),
\end{equation}
where 
\begin{equation}
\label{eq:loss_sal}
\Loss_{\text{sal}}\left(\tilde{\bbi}\right) = \left\|\bbm\circ\left( S(\tilde{\bbi})- \bbt\right)\right\|^2 \;\;\text{and}\;\; \Loss_{\text{sim}}\left(\tilde{\bbi}\right) = \left\|\left(1-\bbm\right)\circ\left(\tilde{\bbi} - \bbi\right)\right\|^2,
\end{equation}
with a saliency model $S(\cdot)$ that predicts a spatial map (per-pixel value in the range of $[0,1]$), and a target saliency map $\bbt$. $\|\cdot\|$ and $\circ$ represent the $L_2$ norm and the Hadamard product, respectively.

We typically use $\bbt \equiv 0$ to reduce the saliency within the region of interest. However, $\bbt$ can be an arbitrary map, so saliency can be increased (e.g., by setting $\bbt \equiv 1$) or set to specific values in the range $[0,1]$, as we show in a couple of examples in the paper and in the supplementary material.




Since existing saliency models are trained on natural images, a naive manipulation of the image pixels guided by Eq.~\eqref{main_naive_obj} can easily converge into "out-of-distribution" outputs. For instance, if additive noise is applied to the pixels within $\bbm$ and optimized with $\bbt\equiv 0$, the output may exhibit salient regions which are interpreted as non-salient by the model, as shown in Figure~\ref{fig:adversarial_attack}.

In order to prevent convergence into the vacant regions of the saliency model, we constrain the solution space of $\tilde{\bbi}$ by substituting $\tilde{\bbi} = O_{\theta}(\bbi)$ in Eq.~\eqref{main_naive_obj}, where $O_{\theta}$ is a pre-defined differentiable operator with a set of parameters $\theta$ that are used as the optimization variables. The constrained objective function can be written as
\begin{equation}
\label{eq:main_obj}
\argmin_{\theta} \; \Loss_{\text{sal}}\left(O_{\theta}(\bbi)\right) + \beta \Loss_{\text{sim}}\left(O_{\theta}(\bbi)\right) + \gamma \Gamma (\theta),
\end{equation}
where $\Gamma(\cdot)$ is a regularization function that is applied to $\theta$, with weight $\gamma$. 

Constraints imposed by using specific operators guarantee that the manipulated images remain within the valid input domain of the saliency model where its predictive power is useful. We next show how different operators $O_{\theta}$ can yield different effects, hand-crafted or learned, that comply with cognitive perception principles~\cite{frintrop2010computational,wolfe2017five}.



Note that the results presented in the paper are achieved by a gradient decent optimization, however, the framework can be converted to a per-operator feed forward network, once trained on scale, as done in other domains such as image style transfer \cite{gatys2016image,johnson2016perceptual}.

\begin{figure}
\centering
\vspace{-.2in}
\includegraphics[width=.75\columnwidth]{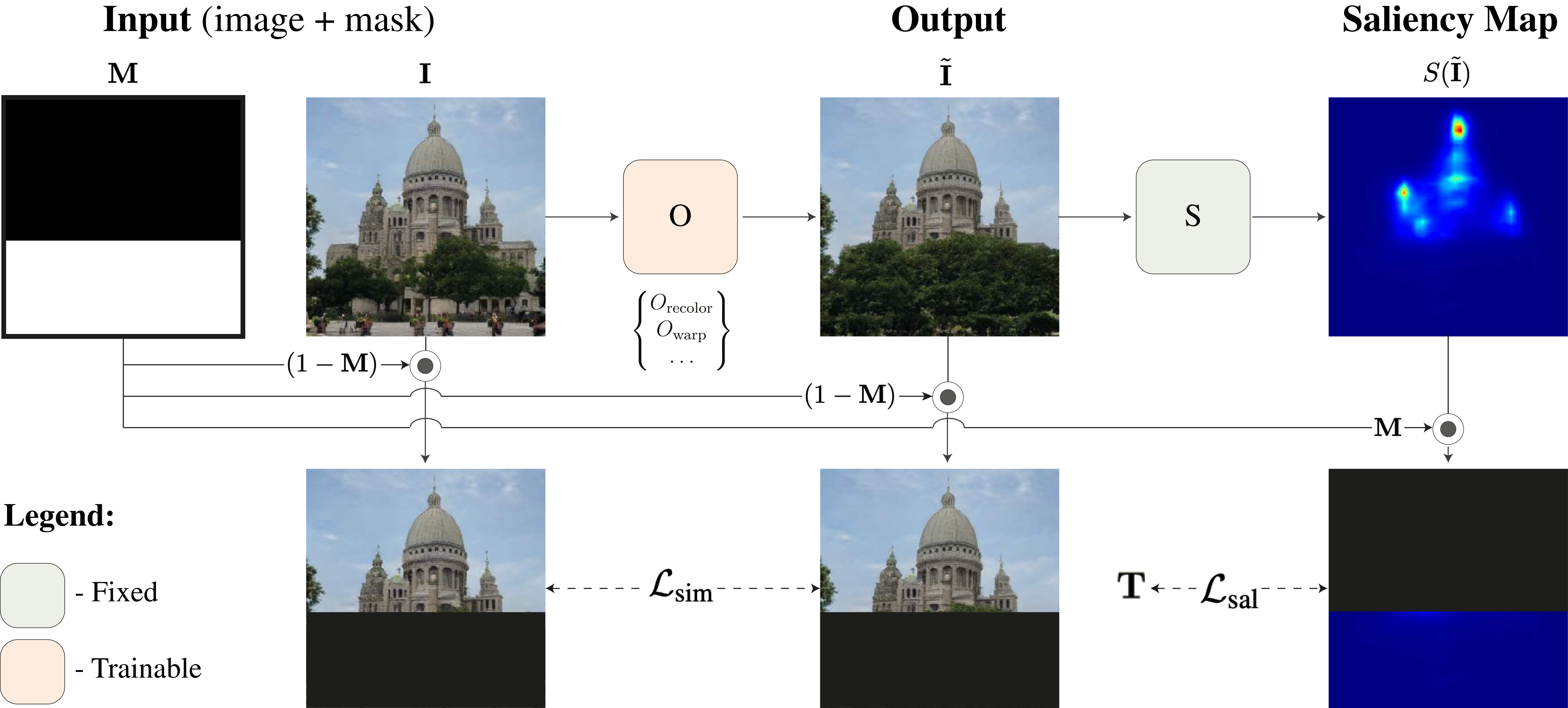}
\vspace{-.1in}
\caption{Our framework. Given an input image $\bbi$, a region of interest mask $\bbm$, and an operator $O \in  \{O_{\text{recolor}}, O_{\text{warp}}, O_{\text {GAN}},\dots\}$. Our approach generates an image with high-fidelity to the input image outside of the mask ($\Loss_{\text{sim}}$), and with reduced saliency inside it ($\Loss_{\text{sal}}$). The target saliency is typically selected to be $\bbt\equiv 0$.
\afterfigure}
\label{fig:prog_masks}
\end{figure}

\textbf{Recolorization -} We first aim at solving a re-colorization task for our purpose, namely, maintaining the luminosity of the region of interest while modifying its chromatic values (`ab' components in the CIELab color representation) in order to reduce saliency. Here, $O_{\theta}$ is a recolor operator that applies a per-pixel affine transform on the `ab' channels of the input image.
The map is represented by a grid $\theta\in\mathbb{R}^{B\times B \times 6}$, that contains $B\times B$ affine transforms. Following the idea of Bilateral Guided Upsampling \cite{chen2016bilateral}, we apply the map to the image in two differentiable steps. In the first step, we extract the affine transforms correspond to each pixel by querying the grid with the `ab' value of the pixels. For example, a pixel with chromatic values $(a,b)$, that lies in the $(i,j)$-th bin, yields the following affine transform
\begin{equation}
    \bbt_{(a,b)} = w_0(a,b)\theta(i,j) + w_1(a,b)\theta(i+1,j) + w_2(a,b)\theta(i,j+1) + w_3(a,b)\theta(i+1,j+1),
\end{equation}
where $w_i(a,b),\;i\in\{0,1,2,3\}$ are bilinear weights that are dictated by the relative position of $(a,b)$ within the bin, and $\bbt_{(a,b)}\in\mathbb{R}^{6}$ is a vector that can be reshaped into the rotation $\bba\in\mathbb{R}^{2\times 2}$ and translation $\bb\in\mathbb{R}^{2}$ parts of the affine transform.
The extracted transformation is applied to the pixel via 
     $\begin{pmatrix}
        a^{\prime}  &
        b^{\prime}
    \end{pmatrix} 
    = \begin{pmatrix}
        a  &
        b
    \end{pmatrix}\bba + \bb$,
where $(a^{\prime},b^{\prime})$ are the output chromatic values. 
In addition, to encourage color changes to be piecewise smooth, we add a smoothness term in the form of an isotropic total variation (TV) loss, $\Gamma (\theta) = \|\nabla_a \theta\|_1 + \|\nabla_b \theta\|_1,$ where $\nabla_a$ and $\nabla_b$ represent the gradients of the grid with respect to the chroma axes $a$ and $b$, respectively. 

\begin{figure}
\centering
\vspace{-.2in}
\includegraphics[width=\columnwidth]{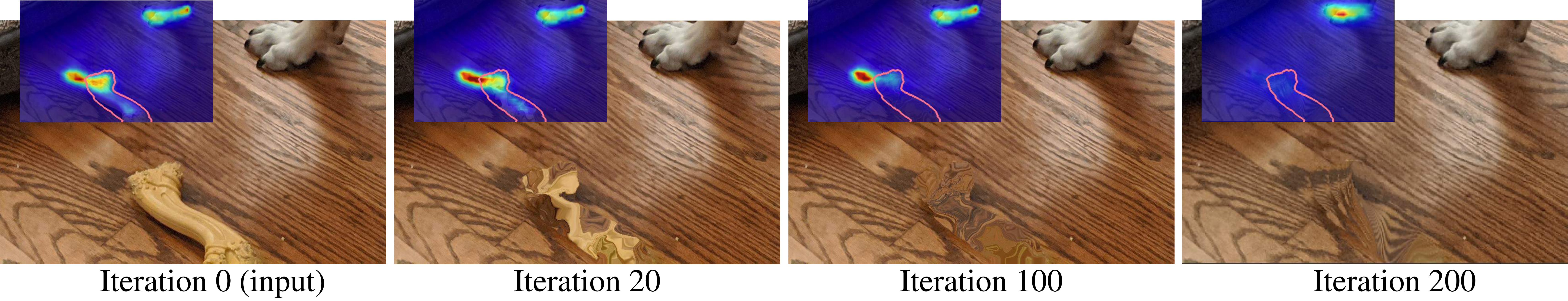}
\vspace{-.2in}
\caption{Saliency driven image warping. Our optimization framework gradually removes the distracting object by covering it with nearby pixels. Texture mismatch results in high saliency, thus, the saliency model guides the warp operator towards a seamless completion of the region. \afterfigure}
\label{fig:warp_seq}
\end{figure}

\textbf{Warping -} We next find a 2D warping field that modifies the saliency of the target region once applied to the image. Here $O_{\theta}$ is a warp operator, represented by a sparse set of control points $\theta$ that are uniformly populated over the image grid. Each control point contains a 2D coordinate that indicates its displacement to the corresponding source pixel. The warp is accomplished in 2 steps. We first upsample the low-resolution grid $\theta$ to the full image size using bilinear interpolation to get the upsampled warp field $\bbw$, then we apply $\bbw$ to the source image. The output value of each pixel is computed by
\begin{equation}
     \tilde{\bbi}(\tilde{i},\tilde{j}) = w_0(\tilde{i},\tilde{j})\bbi(\tilde{i},\tilde{j}) + w_1(\tilde{i},\tilde{j})\bbi(\tilde{i}+1,\tilde{j}) + w_2(\tilde{i},\tilde{j})\bbi(\tilde{i},\tilde{j}+1) + w_3(\tilde{i},\tilde{j})\bbi(\tilde{i}+1,\tilde{j}+1),
\end{equation}
where $(\tilde{i},\tilde{j}) = \lfloor \bbw(i,j) + (i,j) \rfloor$, and $w_i,\;i\in\{0,1,2,3\}$ are bilinear weights, that are dictated by the relative position of $(\tilde{i},\tilde{j})$ within the bin.
Due to the differentiability of the operators, the gradients can be backpropagated through this chain, enabling calculation of the optimal warping field w.r.t~\eqref{eq:main_obj}. In addition, in order to enable better propagation of pixels warped from the exterior region into the interior region of the mask, in each iteration the input image is updated by the warped image $\tilde{\bbi} \rightarrow \bbi$.
A similar smoothness term to one added to the recolor operator is applied to the warping field.
Our results demonstrate that the warp operator tends to remove the object, as it solves an image inpainting problem under unsupervised setting, namely, replacing the foreground object with a natural completion of the background with no explicit self-supervision. Unnatural completion of the background, or mismatch in texture, are interpreted as attention grabbing regions by the saliency model as can be seen in Figure~\ref{fig:warp_seq}. 

 
\begin{figure}
\centering
\vspace{-.2in}
\includegraphics[width=\columnwidth]{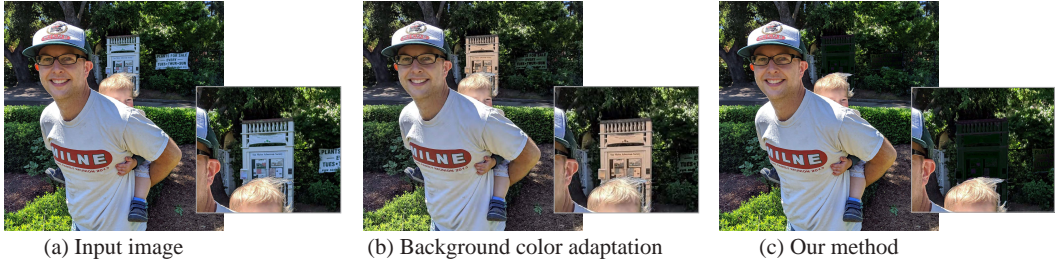}
\vspace{-.2in}
\caption{A comparison against a naive method for adaptation of background colors. (a) The input image, where we wish to reduce the saliency of the sign/post in the back. (b) The result when replacing the chromatic channels with the dominant chromatic values of the surrounding pixels + equalizing the average brightness level with the surrounding pixels by a translation. (c) Our result using the deep conv operator. \afterfigure}
\label{fig:baseline_background_color}
\end{figure}

\textbf{Learning Convolutional Networks} -  We next use an untrained deep convolutional neural network as an image-to-image operator.
The network consists of 5 convolution layers followed by non-linearity (ReLU), where $\theta$ represents the weights of the convolution kernels. Since deep networks may represent a large set of functions, the model can easily converge into an out-of-domain example. Thus, $\Loss_\text{sim}$ plays a key role in maintaining the solution in the valid region of the model. In the first 50 iterations the network weights are optimized to only reconstruct the original image (identity mapping), then the saliency objective is added. 
It can be seen that the network learns to camouflage prominent objects, and blend them with the background~\cite{chu2010camouflage}. Another interesting insight is that the network selects to adapt colors of regions that are associated with the background, even when multiple regions are presented nearby the region of interest (including foreground objects or subjects). Although the network is optimized on a single image (similarly to~\cite{ulyanov2018deep}), the saliency model that was trained on many examples refer background colors to lower saliency, and guides the network to transfer colors of background regions. To demonstrate this point, we calculate a naive baseline which adapts the colors of the surrounding pixels into the marked regions. The chromatic channels were replaced by the most dominant chromatic values of the surrounding pixels, and the brightness is translated such that its average is equal to the average brightness of the surrounding pixels. As can be seen in Figure~\ref{fig:baseline_background_color}, such a naive approach can not distinguish between foreground and background pixel values, while our method can by simply relying on the guidance of the saliency model.


\textbf{StyleGAN as a Natural Image Prior - }
We can further constrain the solution space to the set of natural image patches that can fill the region of interest in a semantically-aware manner. Since this requirement is too general, we incorporate a domain specific pre-trained StyleGAN generator (e.g., human faces, towers, churches), that enables generation of high-quality images from a learned latent distribution, and define $\theta$ to be a latent vector in the $\mathcal{W}$ space~\cite{karras2019style}.  

\begin{figure}
\vspace{-.1in}
\centering
\includegraphics[width=\columnwidth]{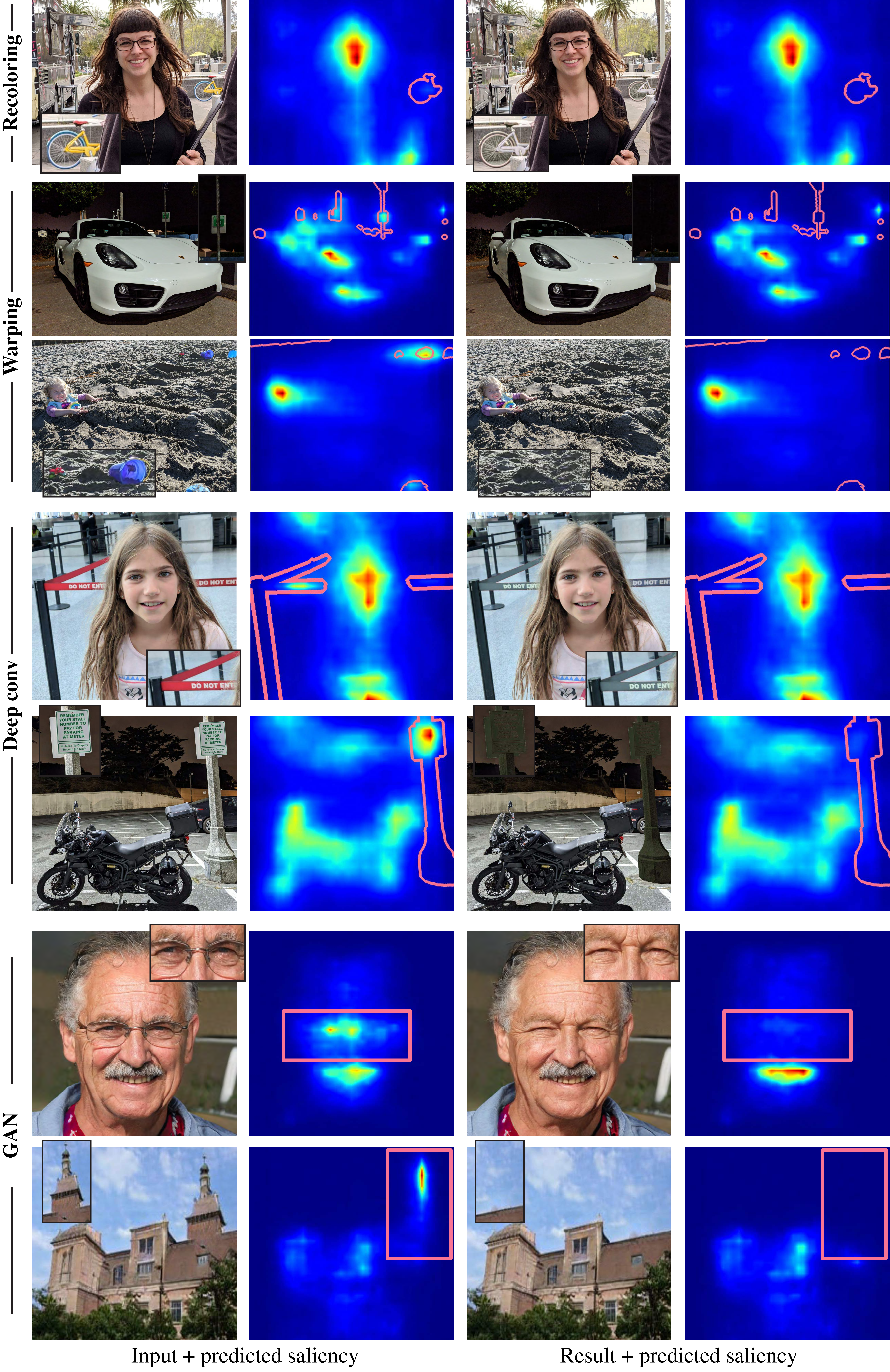}
\vspace{-.2in}
\caption{Additional results of reducing visual distractions, guided by the saliency model with several operators. The region of interest is marked on top of the saliency map (red border) in each example. More results are available in the supplementary material.}
\label{fig:results}
\end{figure}

Given an image $\bbi_{w_0}=G(w_o)$ that was generated by a generator $G$ with a latent code $w_0 \in \mathcal{W}$, 
we initialize $\theta$ to be $\theta_0 = w_0$, and optimize it w.r.t \eqref{eq:main_obj}. To avoid our-of-distribution solutions the output image is restricted to lay in the $\mathcal{W}$ space, by $\tilde{\bbi} = G(\theta)$. The optimization guides the latent code into directions that maintain the details of the image anywhere outside the region of interest, but modify the region's content in a semantically meaningful manner that affects the saliency. For example, in order to reduce the saliency of a structure that contains fine grained details (arcs, poles and windows), the saliency model guides the network to cover the structure by trees. In addition, the model can remove facial accessories such as glasses and to close the eyes of a person (Figure \ref{fig:results}), which comply with cognitive perception principles \cite{fan2018emotional}.

\begin{figure}
\centering
\vspace{-.25in}
\includegraphics[width=.9\columnwidth]{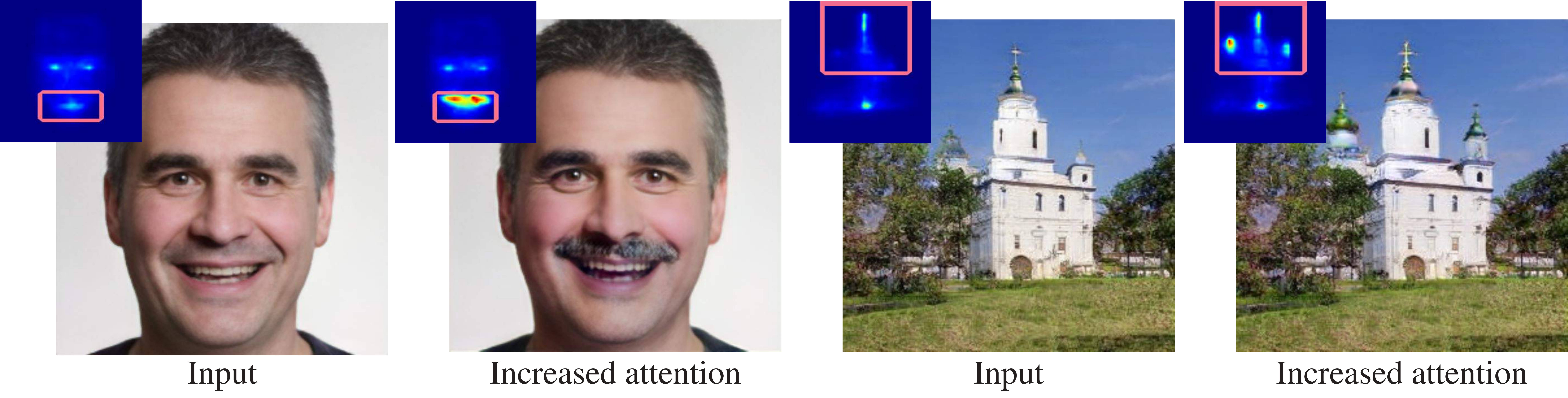}
\vspace{-.1in}
\caption{Saliency \emph{increase} by StyleGAN. For each image pair, the output image (right) was achieved by learning directions in the latent space, such that the saliency of the original image (left) is increased in the region of interest (marked in red on the corresponding saliency map). The found directions are semantically meaningful and natural (adding a moustache and adding prominent domes). \afterfigure}
\label{fig:gan}
\end{figure}

While increasing the saliency of a region is a less-constrained problem that can be solved in various ways with the aforementioned hand-crafted operators (e.g, 'recolor' can modify the colors of the region to be shiny and unnatural, and warp can lead to unnatural attention grabbing distortions), here, the dense latent space of StyleGAN contains a variety of meaningful directions that result in saliency increase. For instance, the saliency model can guide the network to add facial details such as a moustache to increase the saliency in the mouth region, and also add prominent geometric structures such as domes to churches, as shown in figure~\ref{fig:gan}. 

We show semantic editing examples, that are applied to both purely generated images, and examples that were reconstructed from real images using GAN inversion techniques in the supplementary. \ygc{and semantic editing in other space - StyleSpace, as shown in ...}

\section{Results and Experiments} \label{sec:results}
A gallery demonstrating our results with the different operators presented in Section~\ref{sec:method} is depicted in Figure~\ref{fig:results}. More results can be found in the supplementary material. Note that the saliency model guide the operators to mitigate mismatch on color, intensity, texture (spatial frequency), shape, etc., between regions of interest and their surroundings, consistent with existing research on cognitive perception and human visual attention \cite{frintrop2010computational,Itti2007scholarpedia,wolfe2004attributes,wolfe2017five}.

In order to evaluate our method, we collected 120 images and asked professional photography editors to mark regions that draw attention away from the main subjects and reduce the overall user experience~\cite{fried2015finding}.
For the domain-specific GAN approach, we use images from the FFHQ dataset~\cite{karras2019style} and the LSUN dataset~\cite{yu15lsun} for churches and towers. Our framework is implemented in TensorFlow and the parameters of the operators are optimized with the loss term in~\eqref{eq:main_obj} using the Adam optimizer~\cite{kingma2014adam}. More detail can be found in the supplementary material.

\begin{figure}[h]
\centering
\includegraphics[width=0.9\columnwidth]{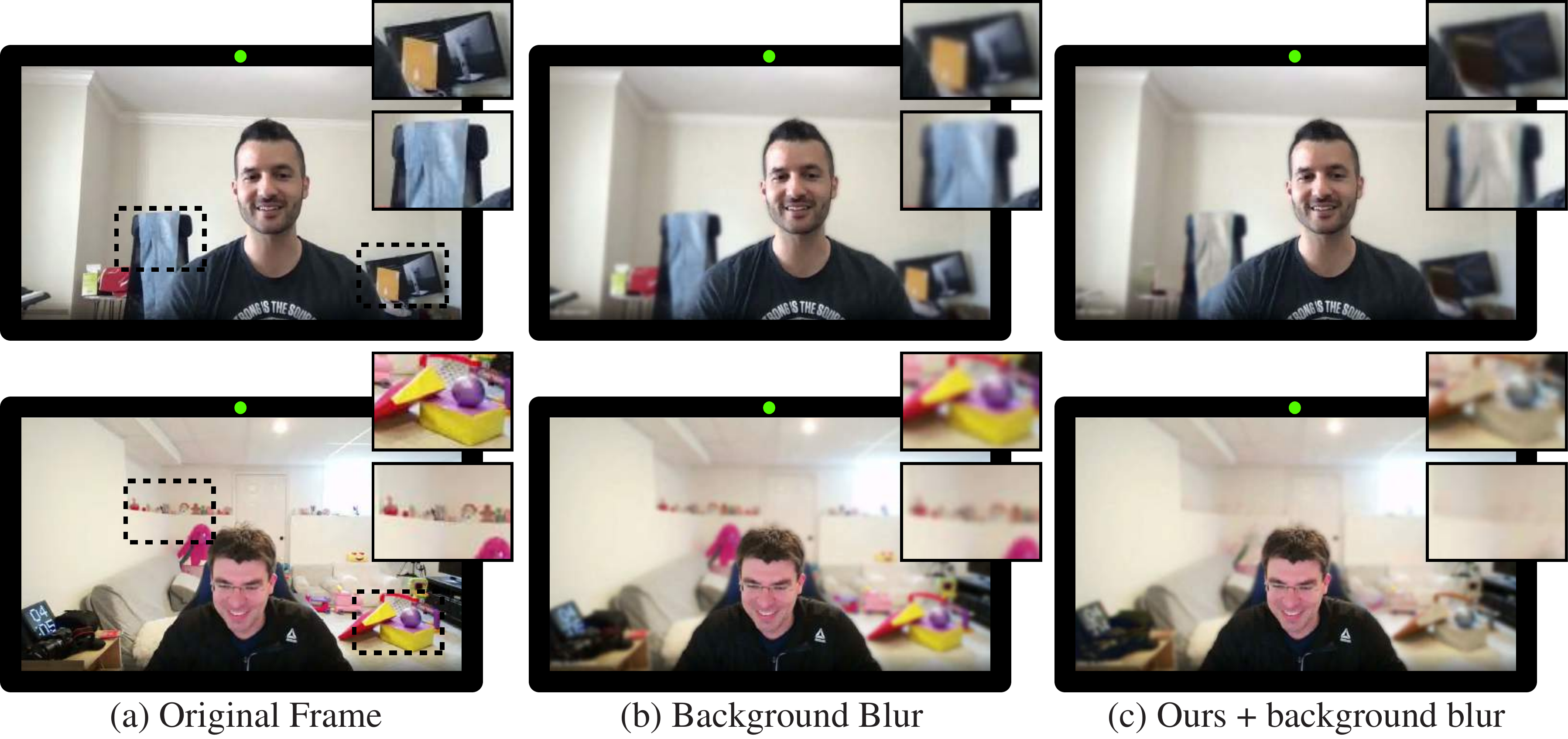}
\vspace{-.1in}
\caption{Reducing distraction in a video conference call (a). Our approach + background blur (c) can reduce visual attention drawn to distracting regions, while maintaining the structural integrity of the subject's environment. Compare with the common background blur effect (b), which leaves colorful, attention-grabbing blobs in the background. See supplementary material for the full video. \afterfigure}
\label{fig:gvc}
\end{figure}

We also demonstrate how our approach can be applied to video conference calls, aiming at reducing background clutter while maintaining the overall appearance of the room or the office. To apply our approach to videos, we segment the regions where the predicted saliency is above a threshold ($t=0.15$). For each distracting region, we apply our different operators and select the one that yields the lowest saliency value within the region and apply the per-distractor parameters to the corresponding regions in all the frames. Figure~\ref{fig:gvc} shows representative frames from the original video, a standard background blur effect, and our effect combined with background blur. It can be seen that our approach selects to inpaint some of the regions using a warp operator while other regions are camouflaged or recolorized. While background blur still includes dominant colorful blobs in the background that may distract from the main speaker, our approach further reduces the attention to the distracting regions while maintaining the overall ``atmosphere'' of the subject's environment.


\begin{figure}
\floatbox[{\capbeside\thisfloatsetup{capbesideposition={right,top},capbesidewidth=4cm}}]{figure}[\FBwidth]
{\caption{Samples of real eye-gaze saliency maps measured in our perceptual study, involving $20$ subjects and $31$ images. Each pair in the first row show an original image (left) with a region of interest on top (red border) and our result (right). The second row depicts the corresponding average eye-gaze maps across participants in the study. }
\label{fig:perceptual_study}}
{\includegraphics[width=9.5cm]{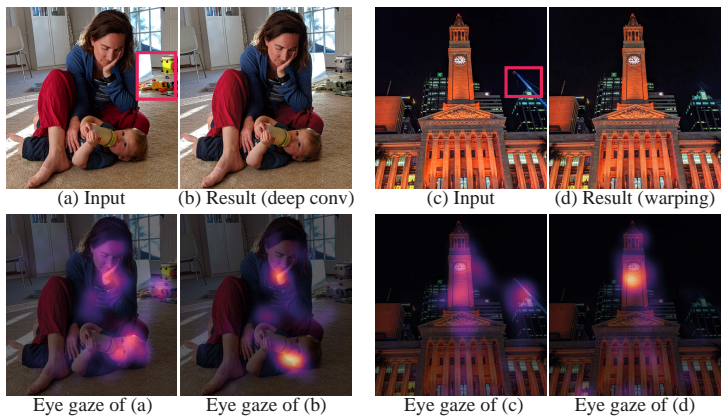}}
\end{figure}

\paragraph{Evaluating Changes in Eye-Gaze}
In order to evaluate the change in eye-gaze that our approach applies to images, we conducted a user study that tracks with high accuracy the eye fixation of $20$ subjects, using the front camera of a smart phone and a dedicated app, as described in \cite{valliappan2020accelerating}\footnote{The gaze data was collected for research purposes only with participants’ explicit consent. In addition, participants were allowed to opt out of the study at any point and request their data to be deleted.}.
The subjects were asked to look at $31$ images, one at a time, where each was presented for 5 seconds followed by a 1 second break. 
In order to ensure that their perception is unbiased, each subject was exposed either to the original image or its modified version, but not to both.
We calculated the gaze saliency map of each image following the common procedure in gaze/saliency study \cite{le2013methods}. 
Figure~\ref{fig:perceptual_study} depicts 2 examples (original and edited) and their average gaze map. It can be seen that the subjects' gaze saliency is reduced in the selected regions (red box), as expected by our approach.
In addition, we compute the mean saliency value within the region, and calculate its average across all the images under each operator. The average reduction, $|\bbm (S_g(\tilde{\bbi}) - S_g(\bbi)) | / |\bbm S_g(\bbi)|$ where $S_g$ is gaze saliency, (per-effect) is reported in Table~\ref{tab:perceptual_study_gaze_saliency} (a). Evidently, our effects successfully reduce the average saliency after the manipulation, demonstrating that our approach guides human attention as expected.
To ensure that our data is statistically significant we ran a statistical test for three features: (i) gaze saliency within the mask, (ii) consecutive gaze duration within the mask, and (iii) first time gaze stays within the mask for more than 50 ms.  We computed the average value (across participants) of each feature and ran a paired samples T-Test (original and edited images as control group and observation group, respectively). The results are shown in Table \ref{tab:perceptual_study_gaze_saliency} (b). In all cases, \emph{p}-value is $<0.003$, implying statistical significance of gaze saliency/duration reduction and first gaze time increase.

\textbf{Realism} Modifying image saliency does not guarantee that the output image is visually plausible, or seems realistic. Hence, we asked 32 users to tell whether a given image looks natural to them. Each user saw 16 images from our dataset, where 4 of them are original and 12 are edited. $85\%$ of the users marked the original images as realistic, while $78\%$ of them marked the same answer to our outputs. The correlation between the numbers implies that our method preserves realism as seem by the users. It is worth noting that in general, researchers must consider the risks of image manipulation techniques, in terms of potential misuse, particularly in the creation of mis/disinformation.


\begin{table}
\centering

\caption{Perceptual study using real gaze - analysis. (a) Reduction of average gaze saliency within the region of interest, per-effect. (b) Paired Samples T-Test (edited images vs. original images) , \emph{p}-value $<0.003$, demonstrating statistical significance. \afterfigure}

\begin{tabular}{c c}
\begin{tabular}{c c c c c}
\toprule
\small \textbf{Recolor} & \small \textbf{Warp} &  \small \textbf{ConvNet} & \small \textbf{GAN} \\
\midrule
  \small 43.1\% &  \small 92.9\% &  \small 53.3\%  &  \small 34.8\%\\
\bottomrule 
\end{tabular} &
\begin{tabular}{c c c}
\toprule
\small \textbf{Gaze saliency} &  \small \textbf{Duration} &  \small \textbf{First gaze} \\
\midrule
\small t=-4.5 &  \small t=-3.3 &  \small t=3.5\\
\small \emph{p}=$6\times 10^{-5}$ &  \emph{p}=\small $2.2\times 10^{-5}$ &  \emph{p}=\small $1.2\times 10^{-3}$\\
\bottomrule
\end{tabular} \\
(a) Gaze saliency reduction & (b) Paired Samples T-Test
\end{tabular}

\label{tab:perceptual_study_gaze_saliency}
\afterfigure
\end{table}

\paragraph{Comparison to state-of-the art}
In order to understand what kind of effects users prefer for the task of saliency reduction, we conducted a study with $32$ participants. The users were asked to look at $16$ images with a marked region of interest, together with two outputs, ours (various effects) and "look-here!", and were asked: “The following two results attempt to draw LESS attention to the region marked in red on the original image. Which one do you like better?”. Table~\ref{tab:comparison}(b) reports the breakdown of user selections between our method and ``look-here!''~\cite{mejjati2020look}. Our results received clear preference for each of the effects, indicating that users in general preferred more aggressive effects to more subtle ones for the purpose of removing distractions.\\
Figure~\ref{fig:look-here} compares our effects visually to ``look-here!'' (more in the supplementary material), and Table~\ref{tab:comparison}(a) reports the percentage of saliency reduction (comparing to the original image) for each of our effects and ``look-here!''. Our method enables a larger reduction in saliency compared to ``look-here!'', as expected from the more dramatic effects we design it for.



\begin{figure}[H]
\centering
\includegraphics[width=\columnwidth]{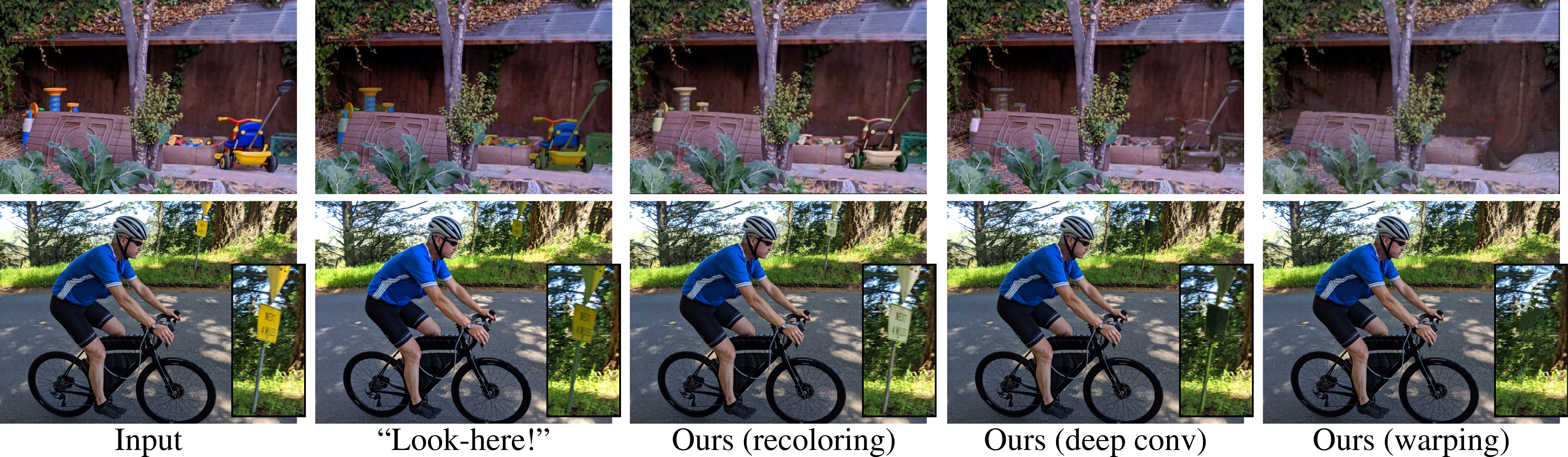}
\caption{Qualitative comparison with ``look-here!''~\cite{mejjati2020look}. More in the supplementary material. In Table~\ref{tab:comparison} we compare numerically with~\cite{mejjati2020look} the effective change to the saliency maps and the users' preferences as found in our user study. \afterfigure}
\label{fig:look-here}
\end{figure}


\begin{table}[H]
\centering
\caption{Comparison of our effects to “look-here!”~\cite{mejjati2020look}. (a) Reduction of average predicted saliency. (b) User study results. We show representative qualitative comparisons to~\cite{mejjati2020look} in Fig.~\ref{fig:look-here}, and more are available in the supplementary material.\afterfigure}

\begin{tabular}{c c}

\begin{tabular}{c c c c c}
\toprule
\small \textbf{Recolor} & \small \textbf{Warp} &  \small \textbf{ConvNet} & \small \textbf{Look-here} \\
\midrule
 \small 43.1\% & \small 92.9\% & \small 53.3\%  & \small 25.8\%\\
\bottomrule 
\end{tabular} &
\begin{tabular}{c c c c}
\toprule
 \small \textbf{"preferred method"} & \small \textbf{Recolor} &  \small \textbf{Warp} &  \small \textbf{ConvNet} \\
\midrule
\small Look-here & \small 31.3\% & \small 9.4\% & \small 18.8\%\\
\small Ours & \small 62.5\% & \small 84.4 & \small 75\%\\
\small "Roughly similar" & \small 6.3\% & \small 6.3\% & \small 6.3\%\\
\bottomrule
\end{tabular} \\
\vspace{-0.05in}
(a) & (b)
\end{tabular}

\label{tab:comparison}
\end{table}
\section{Conclusion}
We introduced a novel framework that utilizes the power of a saliency model trained to predict human eye-gaze, to guide a range of editing effects (e.g., recoloring, inpainting, camouflage, semantic object and attribute editing) that result in meaningful changes to visual attention in images. This is done without any additional training data or direct supervision for the specific editing tasks. In contrast to other methods that use saliency to drive editing which produce subtle changes to image appearance, our method results in drastic, yet realistic edits. We have validated that our results indeed achieve the desired effects on observer’s attention by analysing the changes in real human eye-gaze between the original images and our edited results.

{\footnotesize
{\bf Ethical Considerations.} Our technology focuses on world-positive use cases and applications. Guiding visual attention in images through saliency models has a variety of beneficial and impactful uses, such as removing distractors from photos and video calls, or calling attention to specific areas of a poster or sign to improve the readability and understanding of its content, to name a few. However, we acknowledge the potential for misuse, given the use of generative models to edit images. We emphasize the importance of acting responsibly and taking ownership of synthesized content. To that end, we strive to take special care when sharing images or other material that has been synthesized or modified using these techniques, by clearly indicating the nature and intent of the edits. Finally, we also believe it is imperative to be thoughtful and ethical about the content being generated. We follow these guiding principles in our work.
}

\paragraph{Acknowledgments}
We thank Jon Barron, Phillip Isola, Tali Dekel, Dilip Krishnan and Bill Freeman for their valuable feedback.




{\small
\bibliographystyle{plain}
\bibliography{references.bib}
}

\end{document}